\documentclass[runningheads]{llncs}
\usepackage[T1]{fontenc}
\usepackage{graphicx}
\usepackage{booktabs}
\usepackage{listings}
\usepackage{graphicx}
\usepackage{subfigure}
\usepackage{bbding}
\usepackage{array}

\usepackage{caption}
\usepackage{ragged2e}
\usepackage{caption}
\DeclareCaptionLabelFormat{bf}{\textbf{#1}~\textbf{#2}}
\usepackage{ragged2e}
\usepackage{caption}
\usepackage{xcolor}
\usepackage[table]{xcolor}

\usepackage{multicol}
\usepackage{multirow}
\usepackage{xspace}
\makeatletter
\DeclareRobustCommand\onedot{\futurelet\@let@token\@onedot}
\def\@onedot{\ifx\@let@token.\else.\null\fi\xspace}

\def\etal{\emph{et al}\onedot}
\makeatother

\makeatletter
\newcommand{\thickhline}{%
\noalign {\ifnum 0=`}\fi \hrule height 1pt
\futurelet \reserved@a \@xhline
}
\makeatother

\usepackage{amsmath,amsfonts,bm}

\def\eqref#1{equation~\ref{#1}}

\def\1{\bm{1}}

\def\mH{{\bm{H}}}

\def\mV{{\bm{V}}}
\def\mW{{\bm{W}}}
\def\mX{{\bm{X}}}

\def\mZ{{\bm{Z}}}

\DeclareMathAlphabet{\mathsfit}{\encodingdefault}{\sfdefault}{m}{sl}
\SetMathAlphabet{\mathsfit}{bold}{\encodingdefault}{\sfdefault}{bx}{n}

\usepackage{orcidlink}

\usepackage{marvosym} 

\usepackage{hyperref}

\usepackage{pifont}
\definecolor{softgreen}{RGB}{46,125,50}
\definecolor{softred}{RGB}{198,40,40}
\newcommand{\cmark}{\textcolor{softgreen}{\ding{51}}}
\newcommand{\xmark}{\textcolor{softred}{\ding{55}}}

\begin{document}

\title{Spatial-Temporal Decoupled Adapter for Micro-gesture Online Recognition}

\author{Xucheng Shen\inst{1}\orcidlink{0009-0009-8586-8274} \and
Kun Li\inst{2}\textsuperscript{(\Letter)}\orcidlink{0000-0001-5083-2145} \and 
Fei Wang\inst{1,3,4}\orcidlink{0009-0004-1142-6434}
\and
Wei Qian\inst{1}\orcidlink{0009-0007-9467-6296} \and 
Jin Jiang\inst{2}\orcidlink{0000-0003-4299-6259} 
\and Dan Guo\inst{1,3}\orcidlink{0000-0003-2594-254X}
}

\authorrunning{X. Shen et al.}

\institute{
Hefei University of Technology, Hefei, China \and
United Arab Emirates University, Al Ain, United Arab Emirates
\email{kunli.hfut@gmail.com}
\and Institute of Artificial Intelligence, Hefei Comprehensive National Science Center, Hefei, China \and
Anhui Evolution Technology Co., Ltd., Hefei, China 
}

\maketitle 

\begin{abstract}
Micro-gesture online recognition aims to temporally localize and classify subtle gestures in untrimmed videos. Owing to their extremely short duration, low motion amplitude, and ambiguous visual cues, capturing discriminative spatiotemporal representations remains highly challenging. Existing parameter-efficient adapters typically employ a single branch to model spatial and temporal cues jointly, which may fail to capture the fine-grained patterns of micro-gestures. To address this limitation, we propose a Spatial-Temporal Decoupled Adapter that decomposes video adaptation into independent temporal and spatial branches via lightweight depthwise convolutions. In addition, to alleviate the long-tailed class distribution inherent in the benchmark dataset, we introduce an Adaptive Soft Balanced Augmentation method, which dynamically allocates augmentation intensity based on class rarity and learning difficulty, without manual thresholds. Our method achieves an F1 score of 0.43808, ranking 1st in Track 2 of the 4th EI-MiGA-IJCAI Challenge.

\end{abstract}

\section{Introduction}
Micro-gestures~\cite{chen2023smg,guo2024mac,liu2021imigue} are subtle and frequently unconscious body movements, primarily involving the hands, fingers, and arms, that arise spontaneously in the course of interpersonal communication. 
These inconspicuous behaviors provide valuable insights into human psychological states and emotional conditions, making them increasingly important for affective computing~\cite{gao2026identity,qian2024cluster,qian2025physdiff,wang2026gait}, psychological assessment~\cite{chen2019analyze,qian2024dual,qian2026freqphys}, and intelligent human-computer interaction~\cite{liu2025survey}.

As a highly challenging task, micro-gesture online recognition requires both temporal localization and category classification of micro-gesture instances in untrimmed videos. Compared to traditional action recognition~\cite{hao2022attention,hao2022group} or temporal action detection~\cite{lin2019bmn,zhang2022actionformer}, this task places greater emphasis on distinguishing between fine-grained micro-gesture categories and precisely identifying the start and end times of each instance. The core difficulty lies in capturing subtle differences among micro-gestures to distinguish fine-grained categories and accurately determine the start and end timestamps of action segments. 

Traditional temporal action detection methods~\cite{lin2019bmn,liu2020tsi,zhao2021video} typically rely on a two-stage paradigm, where video features are extracted offline and subsequently used to train a detector. Such a design prevents the backbone from being optimized by the detection loss, potentially limiting representation learning.
End-to-end training overcomes this by jointly learning the backbone and detection head, which theoretically yields stronger representations. However, fully fine-tuning a large model like VideoMAE‑g requires extremely high GPU memory that common hardware cannot afford.

Recent advances~\cite{agrawal2025scaling,liu2024end} have explored parameter-efficient fine-tuning of pretrained vision encoders for temporal action detection, achieving an effective balance between performance and computational efficiency. A common strategy is to insert lightweight adapter modules into frozen backbone layers, enabling the pretrained model to efficiently adapt to the downstream detection task with minimal additional parameters. Most existing designs adopt a single-branch structure that processes spatial and temporal information within a unified pathway, which may limit the model's ability to independently capture local spatial patterns and long-range temporal dynamics. To address this issue, we propose a Spatial-Temporal Decoupled Adapter that decouples spatial and temporal modeling into parallel streams, allowing each branch to specialize in its respective dimension before merging their complementary representations.

In addition to the challenge of modeling subtle spatio-temporal cues, micro-gesture recognition also suffers from severe data imbalance. As the official benchmark of the EI-MiGA-IJCAI Challenge Track 2, the SMG dataset~\cite{chen2023smg} consists of 40 untrimmed videos spanning 16 micro-gesture categories and one non-micro-gesture category, with a highly skewed class distribution where several categories contain significantly fewer samples than the majority classes. To alleviate this long-tail issue, we introduce an Adaptive Soft Balanced Augmentation method, which dynamically adjusts augmentation intensity for each class based on effective sample counts and learning difficulty, without relying on manual thresholds.

In summary, the contributions of our framework are as follows: 
\begin{itemize}
\item We propose a Spatial-Temporal Decoupled Adapter for parameter-efficient fine-tuning, enabling enhanced fine-grained spatio-temporal representation learning for micro-gesture recognition.

\item We introduce an Adaptive Soft Balanced Augmentation method to improve robustness against long-tail class imbalance by adaptively determining augmentation intensity according to class rarity and learning difficulty.

\item Experiments on the SMG dataset demonstrate the effectiveness of both components for micro-gesture online recognition.
\end{itemize}

\section{Related Work}
\subsection{Micro-Gesture Analysis}
Micro-gesture analysis~\cite{chen2023smg,liu2021imigue}, including micro-gesture recognition~\cite{chen2024prototype,gu2025mm,li2023joint,shang2025cross}, micro-gesture online detection~\cite{liu2025online,liu2024micro}, and behavior-based emotion understanding~\cite{xia2025hybrid}, has attracted increasing research attention with the development of dedicated datasets~\cite{guo2024benchmarking,li2026bench,li2025mmad,wang2026imigue}, advanced methods~\cite{gu2025motion,li2025prototypical,liu2026self}, and competitions~\cite{chen20253rd,guo2024mac,li2025mac}. 
The iMiGUE~\cite{liu2021imigue} is the first public dataset for micro-gesture understanding and emotion analysis, containing over 18,000 video clips across 32 categories. SMG ~\cite{chen2023smg} focuses on spontaneous micro-gestures under psychological stress with 16 MG categories and one non-MG category. MA-52~\cite{guo2024benchmarking} extends to 52 micro-action categories, covering fine-grained natural actions in daily communication. MMA-52~\cite{li2025mmad} introduces multi-label annotations to better reflect action co-occurrence in real-world scenarios. These datasets establish standardized benchmarks for research on fine-grained non-verbal behavior.

Micro-gesture online recognition is challenging due to subtle motion patterns and sparse temporal distribution. Early methods adopted skeleton-based approaches: Guo~\etal~\cite{guo2023micro} combined GCNs with multi-scale Transformers, leveraging the structural prior of human joints to capture fine-grained spatial dependencies, though skeleton extraction errors in occluded scenarios limited robustness. Wang~\etal~\cite{wang2024micro} constructed a dual-stream architecture processing RGB and skeleton modalities in parallel, fusing appearance and structural cues through cross-modal attention, which demonstrated that complementary modalities can compensate for the ambiguity inherent in single-source inputs.

Recent methods have shifted toward large-scale pretrained video backbones, exploiting their rich generalizable representations to reduce reliance on hand-crafted features. Liu~\etal~\cite{liu2024micro} introduced learnable query points coupled with Mamba blocks, replacing conventional temporal modeling with selective state-space scanning that offers linear complexity while preserving long-range context, achieving competitive results using only RGB input. In MiGA 2025, Liu~\etal~\cite{liu2025online} proposed data augmentation strategies and spatial-temporal attention modules built upon DyFADet with VideoMAEv2-g features. Their augmentation pipeline alleviated the severe class imbalance in micro-gesture datasets, while the attention modules sharpened boundary localization, achieving an F1 score of 0.38027. Meng~\etal~\cite{meng2025online} adopted a two-stage pipeline using a frozen VideoMAE-g encoder and DyFADet, validating that combining pretrained visual representations with dynamic sequence modeling advances micro-gesture understanding in complex real-world settings.

\subsection{Parameter-Efficient Fine-tuning for Video Understanding}

Parameter-efficient fine-tuning has become a popular strategy for adapting large pretrained models to downstream tasks with limited trainable parameters. Existing PEFT methods mainly include prompt tuning~\cite{jia2022visual,li2021prefix}, low-rank adaptation~\cite{hu2022lora}, and adapter-based tuning~\cite{chen2022adaptformer,houlsby2019parameter}. 
Among these approaches, adapter tuning is a widely used PEFT strategy that freezes most parameters of the pretrained backbone and inserts small learnable modules into intermediate layers. During fine-tuning, adapters allow pretrained vision representations to be adapted to downstream tasks with minimal additional parameters while preserving the general knowledge learned during large-scale pretraining, and learning task-specific transformations while keeping most pretrained parameters fixed.

Recent advances~\cite{chen2022adaptformer,pan2022st,yang2023aim} have demonstrated the strong potential of adapter-based PEFT for adapting large pretrained vision models to downstream video understanding tasks. AdaptFormer~\cite{chen2022adaptformer} introduces lightweight adapter modules into Vision Transformers, enabling efficient task adaptation while preserving the generalization capability of pretrained representations. Following this paradigm, several studies adapt PEFT to video understanding by enhancing pretrained backbones with specialized temporal modules. For instance, AIM~\cite{yang2023aim} introduces temporal attention modules into frozen CLIP encoders, whereas ST-Adapter~\cite{pan2022st} augments transformer blocks with lightweight spatio-temporal operators to facilitate video representation learning.

More recently, PEFT has also been explored in temporal action detection. AdaTAD~\cite{liu2024end} introduces lightweight adapters into a frozen VideoMAE backbone, enabling end-to-end optimization while preserving the efficiency advantages of parameter-efficient tuning. Despite their effectiveness, existing adapter-based methods predominantly perform spatial and temporal adaptation within a shared adaptation pathway. Such a design may be insufficient for micro-gesture recognition, where discriminative cues are characterized by subtle appearance variations and weak motion patterns. To better capture these fine-grained spatio-temporal characteristics, we propose a Spatial-Temporal Decoupled Adapter that explicitly separates spatial and temporal adaptation into dedicated branches, allowing each branch to focus on complementary aspects of micro-gesture representation learning.

\subsection{Temporal Action Detection}
Temporal Action Detection~\cite{lin2019bmn,liu2024end,zhang2022actionformer,zhao2023re2tal} aims to localize and classify action instances in untrimmed videos. Existing methods can be broadly categorized into three paradigms: two-stage approaches, single-stage methods, and end-to-end optimization. Early methods typically adopt a two-stage~\cite{lin2019bmn,liu2020tsi,zhao2021video} paradigm, where temporal proposals are first generated and then classified using separate detection heads. While effective, such approaches suffer from information loss due to offline feature extraction and require multiple training stages.

Single-stage TAD methods have emerged to simplify the pipeline and improve efficiency. ActionFormer~\cite{zhang2022actionformer} introduces a multi-scale transformer architecture with local self-attention within each temporal scale and cross-scale feature fusion, enabling direct per-frame classification and boundary regression without explicit proposals. TriDet~\cite{shi2023tridet} enhances boundary precision via a trident-head design, while TemporalMaxer~\cite{tang2023temporalmaxer} demonstrates that simple max-pooling can replace complex temporal attention with competitive accuracy at lower computational cost. DyFADet~\cite{yang2024dyfadet} further leverages dynamic feature aggregation to adaptively fuse multi-scale temporal features based on content-dependent routing.

End-to-end approaches have also gained attention, particularly for scenarios with large pretrained backbones. PointTAD~\cite{tan2022pointtad} adopts a sparse query-based paradigm with learnable keyframe points for multi-label temporal action detection using only RGB input. Building on this idea, AdaTAD~\cite{liu2024end} introduces lightweight temporal adapters into a frozen VideoMAE backbone, jointly optimizing the backbone and a multi-scale detection head in an end-to-end manner. This eliminates the conventional two-step pipeline of offline feature extraction followed by separate detection training, and substantially improves localization quality. AdaTAD++~\cite{agrawal2025scaling} further explores spatial-temporal decoupling for scalable action detection, mainly through separated adapter training and high-resolution inference. Different from it, our method adopts a parallel dual-branch adapter within each backbone block, where spatial and temporal branches are jointly trained and their complementary adaptation features are fused for fine-grained micro-gesture recognition.

\section{Methodology}
\subsection{Problem Definition}
Micro-gesture online recognition can be formulated as follows: given a streaming 
video $\mV \in \mathbb{R}^{T\times H \times W \times 3}$, where $H$ and $W$ 
denote the height and width of each frame and $T$ is the number of frames processed 
so far, its micro-gesture annotations can be represented as 
$\Psi_g = \{\varphi_i = (t_s, t_e, c)\}_{i=1}^{N}$, where $t_s$, $t_e$, and $c$ 
are the start time, end time, and category of micro-gesture instance $\varphi_i$, 
and $N$ is the total number of ground-truth instances. The goal of online recognition 
is to predict a candidate proposal set 
$\Psi_p = \{\hat{\varphi}_i = (\hat{t}_s, \hat{t}_e, \hat{c}, s)\}_{i=1}^{M}$ 
to cover $\Psi_g$ in a causal manner, where $s$ denotes the confidence score.

\subsection{Overall Architecture}
We design a parameter-efficient adaptation framework that preserves the spatial representation capability of pretrained Vision Transformers while incorporating necessary temporal modeling through lightweight modules. Given an input video clip $\mX \in \mathbb{R}^{T \times H \times W \times 3}$, a 3D patch embedding layer first converts it into a token sequence $\mZ \in \mathbb{R}^{N \times C}$, where $C$ is the embedding dimension and $N$ is determined by the spatio-temporal resolution after patchification. This sequence is then fed into a pretrained backbone with positional encodings added to preserve spatio-temporal structure.

\begin{figure}[t]
\centering
\includegraphics[width=1\linewidth]{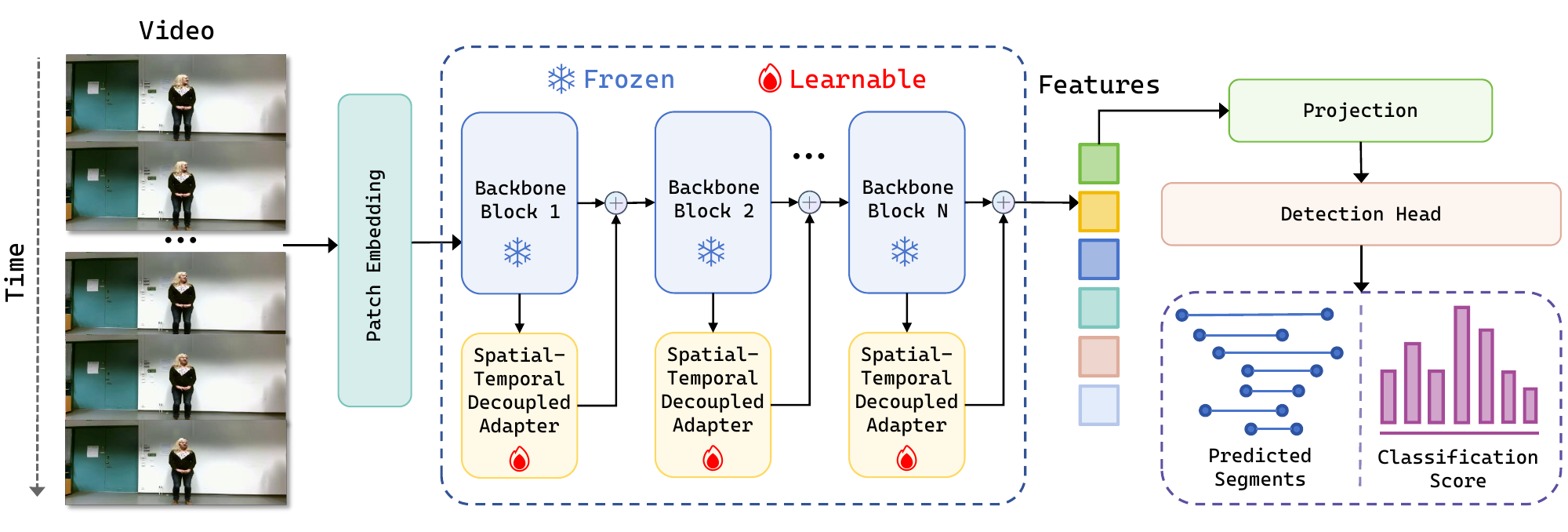}
\caption{Micro-gesture online recognition framework. The input video is patchified and processed by a frozen backbone. Trainable decoupled adapters separately model spatial and temporal information. Aggregated features are fed to the detection head for localization and classification.}
\label{fig:framework}
\end{figure}

To enable task adaptation while retaining pretrained knowledge, we freeze all backbone parameters and insert Spatial-Temporal Decoupled Adapters after each transformer block. As illustrated in Fig.~\ref{fig:framework}, these lightweight bottleneck structures independently model spatial appearance patterns and temporal dynamics, avoiding feature entanglement and improving discriminative ability. After feature extraction, the token sequence is passed through a projection layer while preserving its temporal structure. The projected features are subsequently processed by the detection head to jointly predict temporal boundaries and category labels for micro-gesture instances.

\subsection{Adaptive Soft Balanced Augmentation}

\begin{figure}[t]
\centering
\includegraphics[width=1\linewidth]{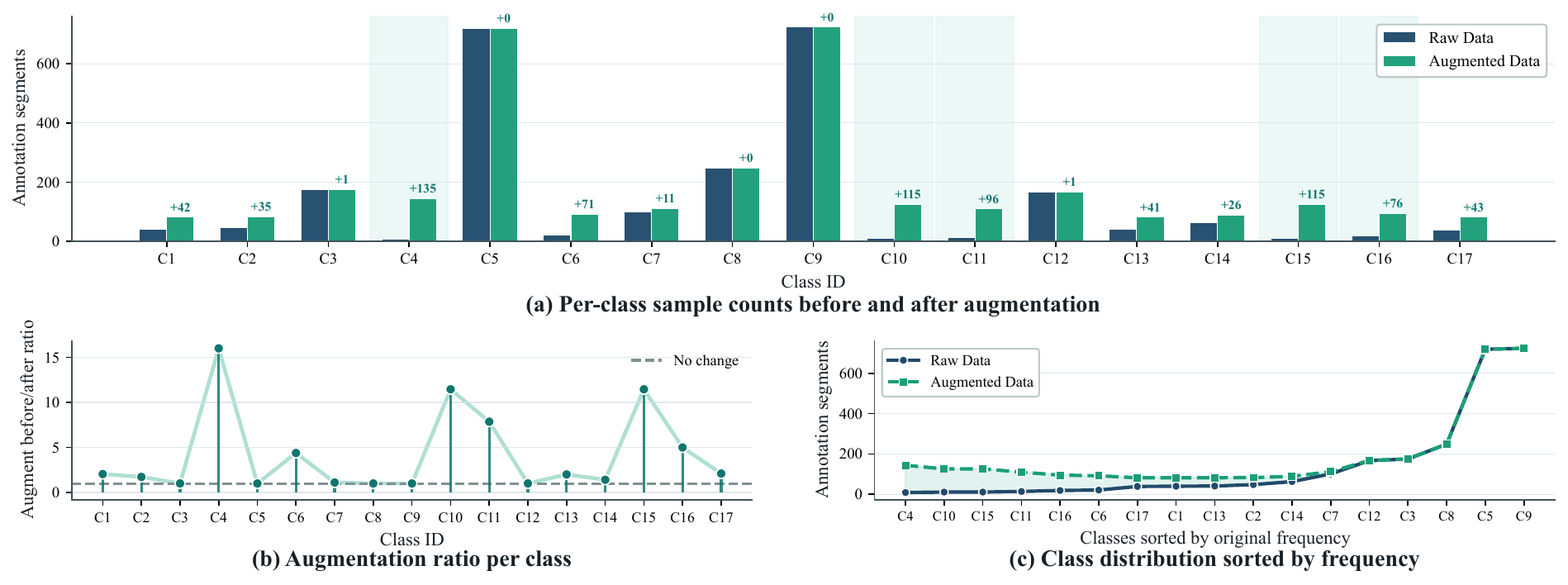}
\caption{Effect of ASBA on the class distribution of the SMG training set. (a) Raw vs augmented annotation segments. ASBA selectively boosts tail categories while keeping head classes nearly unchanged. (b) Class-wise augmentation ratios. Underrepresented classes receive stronger augmentation. (c) Sorted class frequency before and after augmentation. ASBA alleviates long-tailed imbalance while preserving the original data structure.}
\label{fig:asba}
\end{figure}

The SMG dataset exhibits a pronounced long-tailed distribution, where a few dominant categories contain significantly more annotation segments than the majority of tail classes. This causes models to be biased toward head classes and severely degrades recognition performance on rare micro-gesture categories. Conventional oversampling or fixed-threshold augmentation strategies fail to account for the varying degrees of underrepresentation across classes, often leading to overfitting on duplicated tail samples or insufficient augmentation for the most scarce categories.

To address this, we propose Adaptive Soft Balanced Augmentation (ASBA), a data-driven augmentation strategy that dynamically allocates augmentation intensity based on class rarity and learning difficulty, eliminating reliance on manually specified thresholds. 
Inspired by the effective number of samples proposed by Cui \etal~\cite{cui2019class}, which captures the diminishing marginal benefit of additional data points through a geometric data overlap model, we compute an \emph{effective sample count} for each class $c$:
\begin{equation}
E_c = \frac{1 - \beta^{n_c}}{1 - \beta},
\end{equation}
where $n_c$ denotes the original sample count of class $c$, and $\beta$ is a decay coefficient that controls the diminishing marginal contribution of additional samples.

Based on the effective sample count, we define a rarity weight:
\begin{equation}
w_r(c)=\frac{1/E_c}{\max_k(1/E_k)},
\end{equation}
this weight assigns larger values to underrepresented classes (Fig.~\ref{fig:asba}).

To further account for class imbalance, we introduce a difficulty weight:
\begin{equation}
w_d(c)=\frac{N_{\text{total}}-n_c}{N_{\text{total}}},
\end{equation}
where $N_{\text{total}}=\sum_c n_c$ denotes the total number of samples across all classes. The rarity and difficulty weights are then linearly combined into a joint adaptive weight:
\begin{equation}
w_c = \lambda \cdot w_r(c) + (1-\lambda) \cdot w_d(c).
\end{equation}

Rather than relying on a manually specified quantile as the balancing target, ASBA computes a data-driven \emph{adaptive target ceiling}:
\begin{equation}
T^{*} = \frac{1}{|\mathcal{C}|} \sum_{c} n_c (1 + w_c).
\end{equation}

For each tail class satisfying $n_c < T^{*}$, the augmented sample count is determined by:
\begin{equation}
\hat{n}_c = \left\lceil n_c + w_c \cdot \alpha \cdot \left(1 - \frac{n_c}{T^{*}}\right)(T^{*} - n_c) \right\rceil,
\label{eq:asba}
\end{equation}
where $\alpha$ controls the overall augmentation intensity. The quadratic gap term ensures that the most underrepresented classes receive proportionally stronger augmentation, while head classes ($n_c \geq T^{*}$) remain unchanged. 

\subsection{Spatial-Temporal Decoupled Adapter}

\begin{figure}[!t]
\centering
\includegraphics[width=1\linewidth]{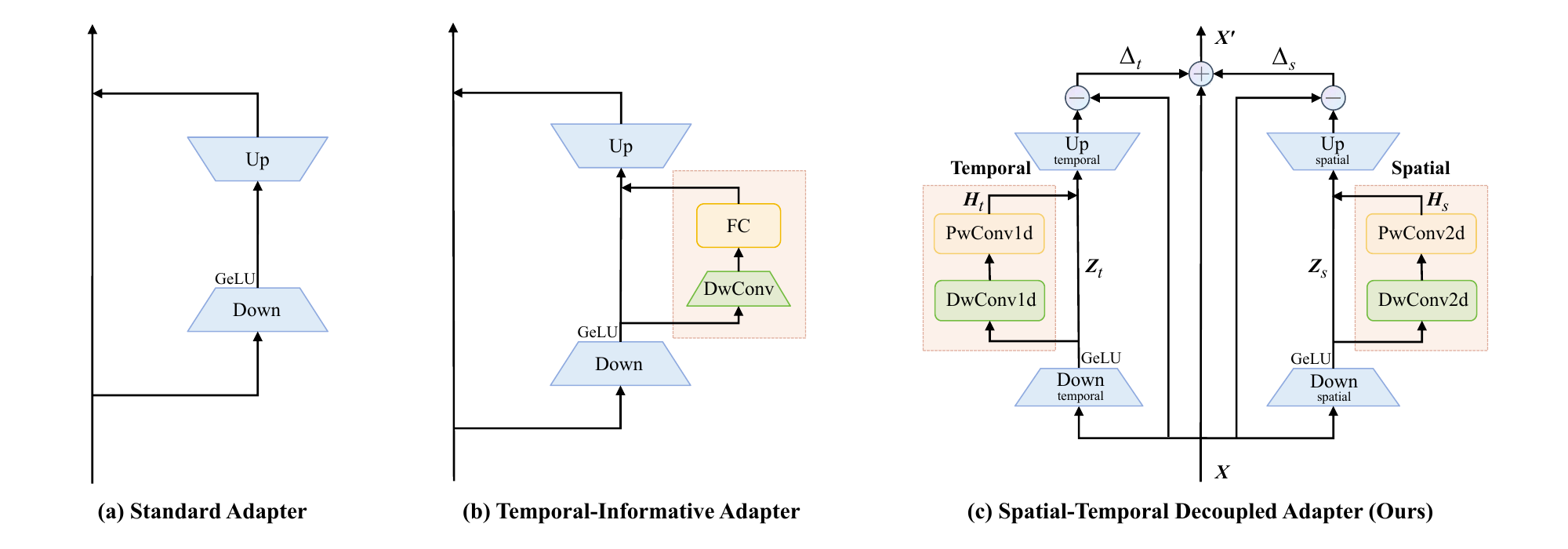}
\caption{Architectural comparison of adapter variants. (a) Standard Adapter: a bottleneck consisting of a down-projection, GeLU activation, and up-projection, with a residual connection. (b) Temporal-Informative Adapter: extends the standard design by inserting a depthwise convolution and a fully connected layer between the down- and up-projections. (c) Spatial-Temporal Decoupled Adapter (Ours): routes the input into parallel temporal and spatial branches, each performing depthwise and pointwise convolutions independently; the deltas from both branches are summed and added back to the input.}
\label{fig:adapter_design}
\end{figure}

Pretrained Vision Transformers encode rich spatial representations but lack temporal modeling capabilities. Existing adapter approaches often apply a single module that jointly processes spatial and temporal information, which may lead to feature entanglement. We argue that spatial appearance patterns and temporal dynamics should be modeled independently to avoid interference and enable more targeted adaptation.

As shown in Fig.~\ref{fig:adapter_design}, Spatial-Temporal Decoupled Adapter consists of two parallel branches inserted after each Transformer block: a Temporal Adapter and a Spatial Adapter. Both branches follow a bottleneck design with down-projection, convolution, and up-projection.

\textbf{Temporal Branch.} 
Given input tokens $\mX \in \mathbb{R}^{B_c \times N \times D}$, where $B_c$ denotes the chunked batch size, $N$ is the number of spatio-temporal tokens in each chunk, and $D$ is the embedding dimension. 
\begin{equation}
\mZ_t = \text{GELU}(\mX \mW_{down}^{t}),
\end{equation}
where $\mW_{down}^{t} \in \mathbb{R}^{D \times D'}$, $D' = \lfloor D \cdot r \rfloor$ with bottleneck ratio $r$. For temporal convolution, the hidden features are then reassembled into to $\mZ_t \in \mathbb{R}^{b \cdot h \cdot w \times D' \times T}$, where $b$ denotes the original batch size, $T$ is the number of temporal positions, and $h$ and $w$ denote the spatial grid dimensions. 

A depthwise 1D convolution is applied along the temporal axis for each spatial position:
\begin{equation}
\mH_t = \text{PWConv1D}(\text{DWConv1D}(\mZ_t)),
\end{equation}
where $\text{DWConv1D}$ denotes depthwise temporal convolution with kernel size $k_t$, and $\text{PWConv1D}$ is a pointwise convolution for channel mixing. The output is reshaped back and projected up:
\begin{equation}
\mX_t = \gamma_t \cdot (\mZ_t + \mH_t) \mW_{up}^{t},
\end{equation}
where $\mW_{up}^{t} \in \mathbb{R}^{D' \times D}$ is zero-initialized for training stability and $\gamma_t$ is a learnable scale parameter.

\textbf{Spatial Branch.} The spatial branch shares the same bottleneck structure but applies depthwise 2D convolutions within each frame:
\begin{equation}
\mH_s = \text{PWConv2D}(\text{DWConv2D}(\mZ_s)),
\end{equation}
where $\mZ_s \in \mathbb{R}^{B_c \cdot t_1 \times D' \times h \times w}$ and $t_1 = N / (h \cdot w)$ denotes the number of temporal tokens per chunk. The spatial convolution captures local spatial patterns such as hand configurations and body part relationships:
\begin{equation}
\mX_s = \gamma_s \cdot (\mZ_s + \mH_s) \mW_{up}^{s},
\end{equation}
where $\mW_{up}^{s} \in \mathbb{R}^{D' \times D}$ is the up-projection matrix of the spatial branch, also zero-initialized for training stability, and $\gamma_s$ is a learnable scale parameter.

\textbf{Parallel Combination.} 
The two branches operate in parallel on the same input $\mX$, and their outputs are combined additively:
\begin{equation}
\begin{aligned}
\mX' &= \mX + (\mX_t - \mX) + (\mX_s - \mX) \\
&= \mX + \Delta_t + \Delta_s,
\end{aligned}
\end{equation}
where $\mX'$ is the adapted output that will be passed to the next Transformer block. The decoupled combination of the two branches ensures that each branch contributes an independent delta, preventing gradient interference between spatial and temporal learning signals.

\section{Experiments}

\subsection{Experimental Setup} 
\textbf{Datasets.} 
The SMG dataset~\cite{chen2023smg} contains 16 MG categories and one non-MG category, collected from 40 subjects.
Following the official cross-subject protocol, 35 subjects are used for training and the remaining 5 for testing. Although both RGB and skeleton modalities are available, our method operates solely on RGB input.

\noindent\textbf{Evaluation Metric.} 
We perform experiments on the SMG dataset and adopt the F1 score and mAP as evaluation metrics. The formula of the F1 score is presented as follows:
\begin{equation}
\text{F1} = \frac{2 \cdot \text{Precision} \cdot \text{Recall}}{\text{Precision} + \text{Recall}},
\label{eq:f1}
\end{equation}
where Precision denotes the fraction of predicted micro-gestures that are correctly classified, and Recall denotes the fraction of ground-truth micro-gestures that are successfully detected. 
This metric captures both the temporal localization accuracy and the classification correctness in a single measure.

\noindent\textbf{Implementation Details.} 
We adopt VideoMAEv2-g~\cite{wang2023videomae} as our video backbone and enhance it with our Spatial-Temporal Decoupled Adapters for feature extraction. The adapter bottleneck ratio is set to 0.25, the temporal kernel size $k_t$ is set to 3, and the spatial kernel size $k_s$ is set to 3. ActionFormer head~\cite{zhang2022actionformer}  is adopted as the detection head for temporal action localization. Videos are processed at the original frame rate of 28 fps with a sliding window of 16 frames and stride of 4 frames. All frames are resized to $160 \times 160$. We adopt the AdamW optimizer with a weight decay of 0.05. The learning rate is initialized to $1 \times 10^{-4}$ and scheduled with cosine annealing over 400 epochs. The batch size is set to 4. For ASBA, we set $\beta = 0.999$, $\lambda = 0.7$, and $\alpha = 1.0$.

\subsection{Experimental Results}

Here, we evaluate our method on the SMG test set and compare with top-performing entries from both the current and previous challenge editions. 
As shown in Table~\ref{tab:Table1}, our method achieves an F1 score of 0.43808, ranking 1st in Track 2 of the 4th EI-MiGA-IJCAI Challenge, surpassing the second-place team by 2.926 percentage points. 
Compared with prior challenge winners, our approach substantially outperforms these methods, demonstrating substantial progress on this task.

\begin{table}[t]
\centering
\caption{The Results of Micro-gesture Online Recognition on the SMG test set. Data is provided by the Kaggle competition page\protect\footnotemark[1].}
\tabcolsep 40pt
\resizebox{\columnwidth}{!}{%
\begin{tabular}{|c|c|c|}
\thickhline
\multicolumn{1}{|c|}{\cellcolor{gray!25}Rank} &
\cellcolor{gray!25}Team &
\cellcolor{gray!25}F1 Score \\
\hline
\multicolumn{1}{|c|}{\cellcolor{gray!10}MiGA'26 1st} &
\cellcolor{gray!10}\textbf{XInsight Lab (Ours)} &
\cellcolor{gray!10}\textbf{0.43808} \\
\hline
MiGA'26 2nd & AIM~\cite{liu2026motion} & 0.40882 \\
\hline
MiGA'26 3rd & XD-L & 0.35559 \\
\hline
MiGA'25 1st & HFUT-VUT~\cite{liu2025online} & 0.38027 \\
\hline
MiGA'25 2nd & Chutian Meng~\cite{meng2025online} & 0.31536 \\
\hline
MiGA'24 1st & NPU-MUCIS~\cite{wang2024micro} & 0.27571 \\
\hline
MiGA'24 2nd & HFUT-VUT~\cite{liu2024micro} & 0.14346 \\
\hline
MiGA'23 1st & NPU-Stanford~\cite{guo2023micro} & 0.14810 \\
\hline
MiGA'23 2nd & HFUT-VUT  & 0.04670 \\ \hline
\end{tabular}}
\label{tab:Table1}
\end{table}

\subsection{Ablation Study}
We conduct ablation studies to evaluate the contribution of each major design choice in our framework.

\textbf{Scaling backbone.} To evaluate the impact of backbone scale, we compare VideoMAE-Small and VideoMAEv2-g in Table~\ref{tab:baseline}.  Scaling up from the small to the giant variant raises the F1 score from 0.35614 to 0.41913, confirming that larger-scale pre-trained representations bring substantial benefits to micro-gesture recognition.

\textbf{Component contributions.} 
To further understand the contribution of each component, we conduct an ablation study in Table~\ref{tab:baseline}. The default single-branch adapter in AdaTAD tends to entangle spatial and temporal features, limiting fine-grained discrimination. To mitigate the class imbalance issue, we apply data augmentation alone and obtain an F1 score improvement from 0.41913 to 0.42450, confirming its effectiveness in alleviating category imbalance. Furthermore, we integrate our proposed Spatial-Temporal Decoupled Adapter together with data augmentation. This combination increases the F1 score to 0.43808, indicating that decoupled spatial-temporal modeling provides a more effective adaptation mechanism for capturing subtle micro-gesture cues.

\textbf{Augmentation strategy.} We evaluate the augmentation strategy on the SMG validation set in Table~\ref{tab:abl_aug}. Training without augmentation yields an average mAP of 25.10\%, highlighting the severe impact of class imbalance. Fixed balancing targets based on mean or median statistics set a target sample size for each class as the mean or median of the class distribution and then augment minority classes up to that target. These targets offer some improvement, but their performance heavily depends on the choice of threshold. In contrast, ASBA achieves 29.05\% average mAP, consistently outperforming both fixed-target baselines across all tIoU levels. This improvement can be attributed to ASBA's mechanism of adjusting augmentation intensity per class according to its effective sample count and learning difficulty, which avoids the over-augmentation or under-augmentation caused by a uniform threshold.

\begin{table}[!t]
\centering
\caption{Ablation results of key modules on the SMG test set.}
\tabcolsep 4pt
\resizebox{\columnwidth}{!}{%
\begin{tabular}{|c|c|c|c||c|}
\hline
\thickhline
\cellcolor{gray!25}Data Augmentation &
\cellcolor{gray!25}Spatial-Temporal Decoupled Adapter &
\cellcolor{gray!25}Method &
\cellcolor{gray!25}Backbone &
\cellcolor{gray!25}F1 score \\
\hline

\xmark & \xmark & AdaTAD~\cite{liu2024end} & VideoMAE-S & 0.35614 \\
\xmark & \xmark & AdaTAD~\cite{liu2024end} & VideoMAE-B & 0.38324 \\
\xmark & \xmark & AdaTAD~\cite{liu2024end} & VideoMAEv2-g & 0.41913 \\
\hline

\cellcolor{gray!10}\cmark &
\cellcolor{gray!10}\xmark &
\cellcolor{gray!10}AdaTAD~\cite{liu2024end} &
\cellcolor{gray!10}VideoMAEv2-g &
\cellcolor{gray!10}0.42450 \\
\cellcolor{gray!10}\cmark &
\cellcolor{gray!10}\cmark &
\cellcolor{gray!10}AdaTAD~\cite{liu2024end} &
\cellcolor{gray!10}VideoMAEv2-g &
\cellcolor{gray!10}\textbf{0.43808} \\
\hline
\end{tabular}%
}
\label{tab:baseline}
\end{table}

\begin{table}[t!]
\centering
\caption{Comparison of augmentation strategies under different tIoU thresholds on the SMG validation set.}
\tabcolsep 4pt
\resizebox{1.0\linewidth}{!}{%
\begin{tabular}{|c|c|c|c|c|c|c|c|c|c|c|}
\hline
\thickhline
\cellcolor{gray!25} &
\multicolumn{9}{c|}{\cellcolor{gray!25}mAP} &
\cellcolor{gray!25} \\
\cline{2-10}

\cellcolor{gray!25}\multirow{-2}{*}{Strategy} &
\cellcolor{gray!25}@0.1 &
\cellcolor{gray!25}@0.2 &
\cellcolor{gray!25}@0.3 &
\cellcolor{gray!25}@0.4 &
\cellcolor{gray!25}@0.5 &
\cellcolor{gray!25}@0.6 &
\cellcolor{gray!25}@0.7 &
\cellcolor{gray!25}@0.8 &
\cellcolor{gray!25}@0.9 &
\cellcolor{gray!25}\multirow{-2}{*}{Avg} \\
\hline

None & 36.16 & 35.95 & 35.13 & 33.48 & 31.33 & 25.61 & 19.61 & 7.22 & 1.43 & 25.10 \\
\hline

Mean & 38.62 & 38.30 & 37.64 & 34.34 & 32.52 & 29.29 & 24.94 & 15.27 & 1.42 & 28.04 \\
Median & 38.55 & 38.23 & 37.54 & 34.28 & 32.45 & 28.98 & 13.25 & 6.21 & 1.34 & 25.65 \\
\hline

\cellcolor{gray!10}\textbf{ASBA (Ours)} &
\cellcolor{gray!10}\textbf{39.04} &
\cellcolor{gray!10}\textbf{38.74} &
\cellcolor{gray!10}\textbf{38.28} &
\cellcolor{gray!10}\textbf{35.75} &
\cellcolor{gray!10}\textbf{33.60} &
\cellcolor{gray!10}\textbf{30.65} &
\cellcolor{gray!10}\textbf{25.55} &
\cellcolor{gray!10}\textbf{17.72} &
\cellcolor{gray!10}\textbf{2.08} &
\cellcolor{gray!10}\textbf{29.05} \\
\hline
\end{tabular}%
}
\label{tab:abl_aug}
\end{table}

\subsection{Error Analysis}

\begin{figure}[t!] 
\centering
\includegraphics[width=1\linewidth]{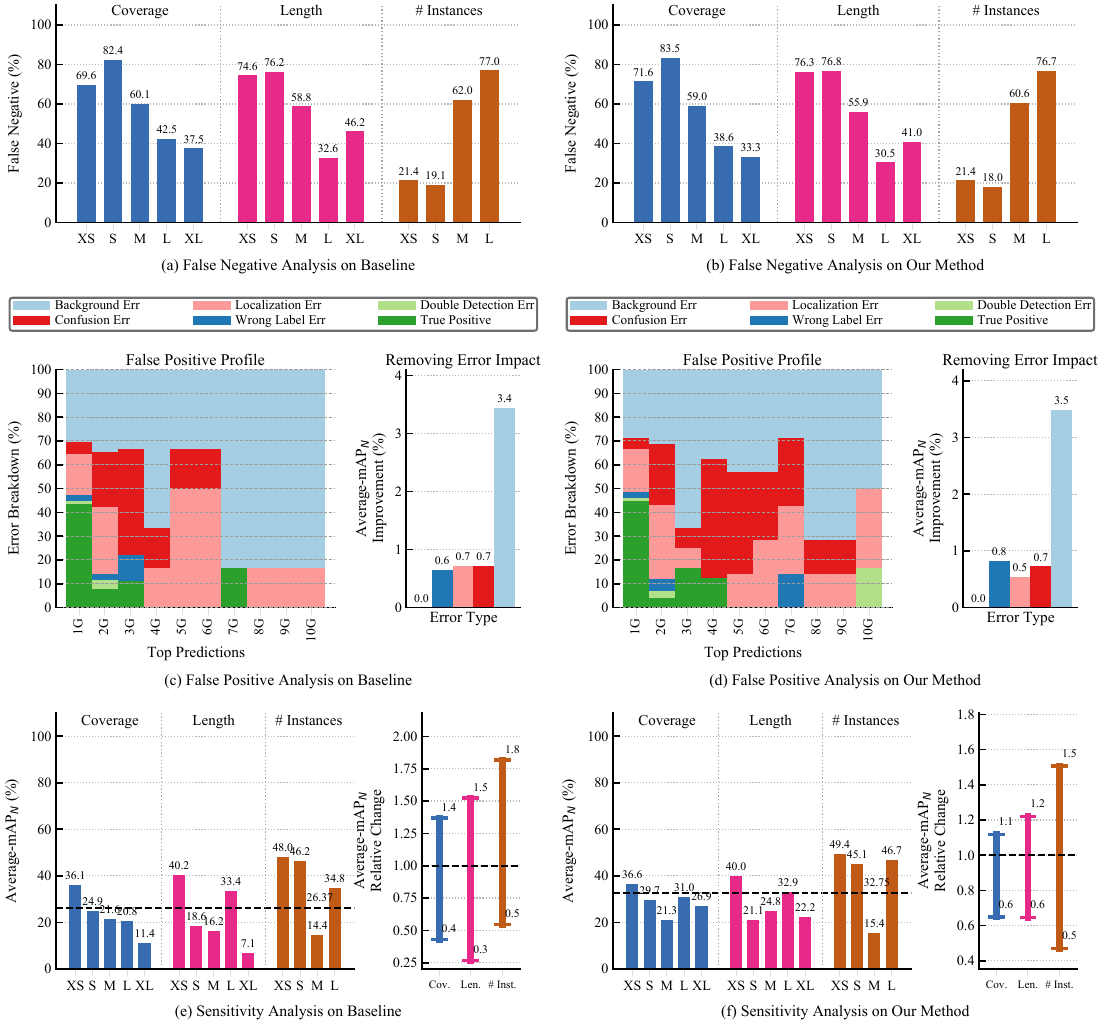}
\caption{Diagnostic evaluation of the baseline and our method on false negatives, false positives, and sensitivity analysis. The baseline adopts the AdaTAD model with VideoMAEv2-g as the backbone without data augmentation.}
\label{fig:diag}
\end{figure}

In addition, we follow the standard practice in temporal action detection by adopting the diagnostic evaluation toolkit proposed by Alwassel~\etal~\cite{alwassel_2018_detad} to analyze model behavior from three perspectives. 
To better match the characteristics of the SMG dataset, we define Coverage groups as [XS, S, M, L, XL] with boundaries [0, 0.001, 0.0015, 0.0025, 0.0045, 1], Length groups as [XS, S, M, L, XL] with boundaries [0, 1, 1.5, 2.5, 4, INF], and Instance groups as [XS, S, M, L] with boundaries [-1, 50, 100, 200, INF].

\textbf{False Negative Analysis.} Fig.~\ref{fig:diag} (a) and (b) compare the missed detection rates at tIoU = 0.5. Our method reduces false negatives primarily in medium-to-large coverage and longer-duration groups, indicating improved detection capability for temporally extended action instances. However, both methods still exhibit high miss rates on very short segments, suggesting that detecting brief micro-gestures remains challenging.

\textbf{False Positive Analysis.} Fig.~\ref{fig:diag} (c) and (d) decompose false positive errors into five categories. Background error dominates the false positive composition in both methods and contributes the largest potential gain if removed. Compared to the baseline, our method slightly increases the true positive proportion in top-ranked predictions and reduces background errors, while localization and confusion errors remain comparable.

\textbf{Sensitivity Analysis.} Fig.~\ref{fig:diag} (e) and (f) evaluate Average-mAP$_N$ across different characteristic groups. Our method improves the overall Average-mAP$_N$ from 26.4\% to 32.7\%, with particularly notable gains in large-coverage, longer-duration, and high-instance-count groups. The relative sensitivity analysis shows that our method achieves more balanced performance across different segment characteristics, with reduced sensitivity to coverage and length variations compared to the baseline.

\section{Conclusion}
In this paper, we present our solution for the Micro-gesture Online Recognition track of the IJCAI 2026 MiGA Challenge. We propose a spatial-temporal-decoupled adapter together with an adaptive soft balanced augmentation strategy to model fine-grained spatial-temporal patterns and alleviate severe category imbalance. Specifically, the proposed adapter introduces parallel temporal and spatial branches into the frozen video backbone, capturing subtle motion dynamics and local appearance cues with few trainable parameters. The augmentation strategy dynamically adjusts augmentation intensity based on class rarity and sample distribution, improving tail-class representation while preserving head classes. Experimental results demonstrate that the proposed method achieves performance improvements by a large margin. 

Although the proposed method achieves notable improvements, the accuracy of temporal localization and boundary detection remains limited. In future work, we plan to incorporate skeleton data, explore multi-modal fusion, and improve the detection head to further enhance temporal localization and recognition performance.

\section*{Acknowledgments}
This work was supported by Anhui Provincial Natural Science Foundation \sloppy (2408085J040), National Key R\&D Program of China (2024YFB3311600), Natural Science Foundation of China (62272144, 72188101), the Major Project of Anhui Provincial Science and Technology Breakthrough Program (202423k09020001), and the New Cornerstone Science Foundation through the XPLORER PRIZE.

\bibliographystyle{splncs04}
\bibliography{refs}

\end{document}